\documentclass[10pt,twocolumn,letterpaper]{article}

\usepackage{iccv}              

\definecolor{iccvblue}{rgb}{0.21,0.49,0.74}
\usepackage[pagebackref,breaklinks,colorlinks,allcolors=iccvblue]{hyperref}
\usepackage{wrapfig} 
\def\paperID{12779} 
\def\confName{ICCV}
\def\confYear{2025}

\title{Towards More Accurate Personalized Image Generation: Addressing Overfitting and Evaluation Bias}

\author{Mingxiao Li\textsuperscript{1}, Tingyu Qu\textsuperscript{1}, Tinne Tuytelaars\textsuperscript{2}, Marie-Francine Moens\textsuperscript{1}\\
\textsuperscript{1} Department of Computer Science, KU Leuven\\
\textsuperscript{2} Department of Electrical Engineering, KU Leuven\\
{\tt\small \{tingyu.qu, mingxiao.li, tinne.tuytelaars, sien.moens\}@kuleuven.be}
}
\begin{document}
\maketitle
\begin{abstract}
Personalized image generation via text prompts has great potential to improve daily life and professional work by facilitating the creation of customized visual content. The aim of image personalization is to create images based on a user-provided subject while maintaining both consistency of the subject and flexibility to accommodate various textual descriptions of that subject. However, current methods face challenges in 
ensuring fidelity to the text prompt 
while not 
overfitting to the training data.
In this work, we introduce a novel training pipeline that incorporates an attractor to filter out distractions in training images, allowing the model to focus on learning an effective representation of the personalized subject.
Moreover, current evaluation methods struggle due to the lack of a dedicated test set. The evaluation set-up typically relies on the training data of the personalization task to compute text-image and image-image similarity scores, which, while useful, tend to overestimate performance. 
Although human evaluations are commonly used as an alternative, they often suffer from bias and inconsistency. To address these issues, we curate a diverse and high-quality test set with well-designed prompts. With this new benchmark, automatic evaluation metrics can 
reliably assess model performance.\footnote{Code and dataset are avaiable at \url{https://github.com/Mingxiao-Li/Towards-More-Accurate-Personalized-Image-Generation}}

\end{abstract}    
\section{Introduction}
\label{sec:intro}
Personalized image generation~\cite{ti,dreambooth,neti} aims to create images that incorporate a user-specified subject or concept in a set of images. The generation adheres to the provided text prompts, while consistently generating the same subject and its appearance.
This technology transforms creative and professional workflows by generating customized visuals from text prompts. Recent advances in text-to-image models enhance quality and adaptability, enabling more accurate, coherent results.
However, despite these advancements, challenges remain.

\begin{figure}[t]  
    \centering
    \includegraphics[width=0.5\textwidth]{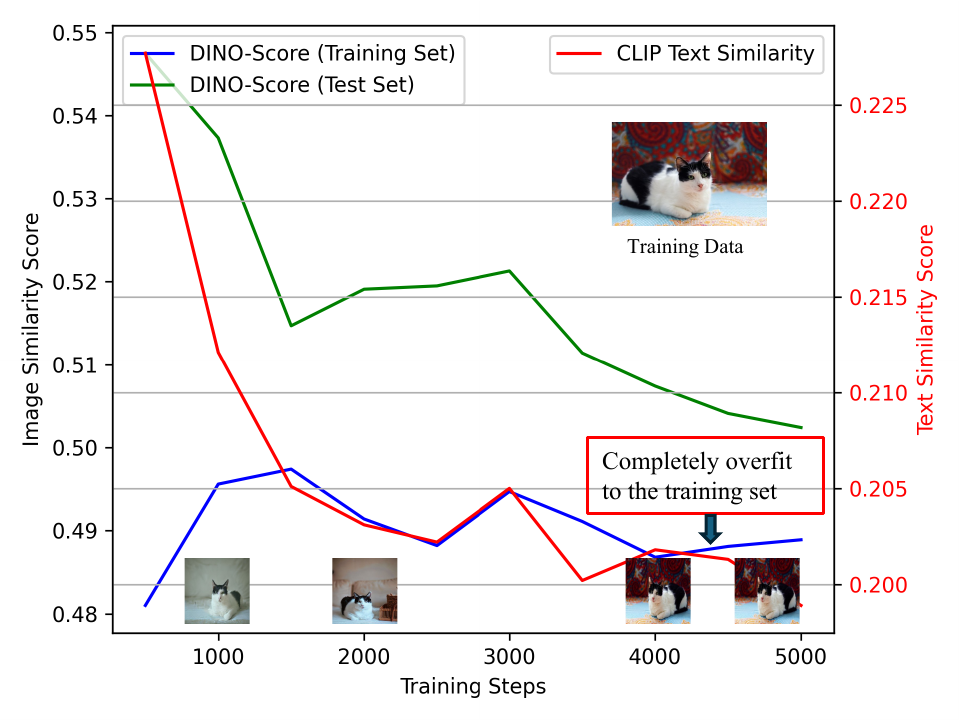}  
    \caption{
    Changes in DINO image-image similarity and CLIP text-image similarity scores for the DreamBooth model across training steps. The DINO score, evaluated on a separate test set, shows a stronger correlation with the CLIP text similarity score. In contrast, the DINO score on the training set remains nearly unchanged, even after the model has fully overfit to the training data.
    }
    \label{fig:overfit}
\end{figure}

One major issue is 
ensuring fidelity to the text prompt 
while not 
overfitting to the training data. Models that overfit may generate images highly similar to those in the training set rather than truly understanding and generalizing the provided subject. Furthermore, the current automatic evaluation framework for personalized image generation is flawed. 
Automatic metrics such as 
CLIP text-image similarity score, CLIP image-image similarity score, and DINO image-image score~\cite{clip, dino, dreambooth}
are widely used to assess both text alignment and subject consistency. However, due to the absence of an independent test set, in previous works these metrics are 
always computed using the training data itself~\cite{ti,dreambooth,neti}. This leads to evaluation biases. Models that overfit the training set can achieve artificially high similarity scores.
For example, in~\cref{fig:overfit}, the higher similarity score of the generated image at 5000 steps misleadingly suggests strong generalization. Addressing such evaluation flaws is crucial for accurately tracking progress in personalized image generation and ensuring that models learn beyond memorizing training data.

The goal of image personalization is to generate images based on a user-provided subject while ensuring 
both subject consistency and adaptability to diverse textual descriptions that regard that same subject. To tackle the aforementioned challenges, we propose a novel training pipeline that seamlessly integrates with existing tuning-based image personalization techniques, such as Textual Inversion~\cite{ti} and NeTI~\cite{neti}. Unlike traditional approaches that solely focus on learning the target subject, our pipeline introduces an attractor, a mechanism specifically designed to filter out distractions in training images. By reducing background noise and emphasizing the primary subject, this method enables more effective subject learning, resulting in higher-quality image generation and improved alignment with user prompts. Additionally, our approach mitigates the risk of overfitting while preserving the model’s ability to generate diverse and highly personalized content.

To address the evaluation bias caused by using training data for automatic metric computation, we curate a diverse and high-quality dataset consisting of both a dedicated training set and a separate test set with carefully designed, detailed prompts. This benchmark enables 
reliable performance assessment by ensuring that evaluation metrics are computed on unseen data. This reduces the risk of overestimated results and mitigates flaws of existing evaluation practices.

In this paper, we present our novel training framework alongside our newly developed benchmark for personalized image generation. We demonstrate that our approach improves subject fidelity while reducing overfitting and enhances the reliability of automatic evaluation metrics. Our contributions can be summarized as follows:

\begin{itemize}
    \item \textbf{A novel training pipeline} that utilizes an attractor to filter distractions, improving the focus on primary subjects in personalized image generation.
    \item \textbf{A curated benchmark dataset} with diverse and high-quality prompts, enabling more reliable automatic evaluation of model performance.
    \item \textbf{Comprehensive analysis} of existing evaluation methods, highlighting their limitations and demonstrating how our benchmark mitigates these issues. 
\end{itemize}
\section{Related Work}
\label{sec:relate-work}
\subsection{Diffusion Models}
Diffusion models are probabilistic generative models that have achieved remarkable success in learning complex data distributions across various domains, including images~\cite{ldm,dalle,chen2023pixartalpha,podell2023sdxl}, videos~\cite{pixels-dance,movie-gen,open-sora,wang2023modelscope,videocrafter1,zhang2024show,kong2024hunyuanvideo,bao2024vidu}, and 3D objects~\cite{poole2022dreamfusion,lin2023magic3d,wang2024prolificdreamer,ye2025dreamreward,tang2023stable}. These models operate through a two-step process: a forward process, which incrementally adds noise to a clean image, and a backward process, which progressively removes the noise to reconstruct the original image. This iterative generative mechanism not only enables high-quality synthesis but also offers significant flexibility for controllable generation, such as image layout control~\cite{zhang2023adding,mou2024t2i} and video motion control~\cite{li2024animate,geng2024motion}. Beyond creative applications, diffusion models have also found widespread use in scientific domains, including protein generation~\cite{prot,trippe2022diffusion,wu2024protein,anand2022protein} and brain visual decoding~\cite{sun2024contrast,sun2023decoding,chen2023seeing,sun2025neuralflix}. However, despite their strong generative capabilities, diffusion models suffer from slow sampling speeds, making them computationally expensive. To address this, researchers have explored methods to accelerate the generative process~\cite{song2020denoising,lu2022dpm,zheng2023dpm}. Meanwhile, another line of research focuses on improving generation quality by mitigating exposure bias~\cite{li2023alleviating,ning2023elucidating}.

\subsection{Personalization}
 To achieve image personalization, existing approaches can be broadly categorized into fine-tuning-based methods and encoder-based methods. Early works, including ~\cite{ti,dreambooth,neti,han2023svdiff,voynov2023p+,kumari2023multi}, primarily focus on fine-tuning techniques. These methods adjust either specific parts or the entire set of weights in the diffusion network while leveraging pseudo-word embeddings to store and represent the target subject effectively. In contrast, encoder-based methods~\cite{huang2024realcustom,li2023blip,pan2023kosmos,patel2024lambda,wei2023elite} eliminate the need for test-time tuning by introducing an additional network—often referred to as an adapter or subject encoder—that learns to encode subject information. While this approach enables faster inference, it requires training on a large-scale personalized dataset, which can be difficult to obtain. In this work, we focus on the fine-tuning-based methods.

\section{Dataset Construction}
\begin{figure*}[htbp]
    \centering
    \includegraphics[width=1.0\textwidth]{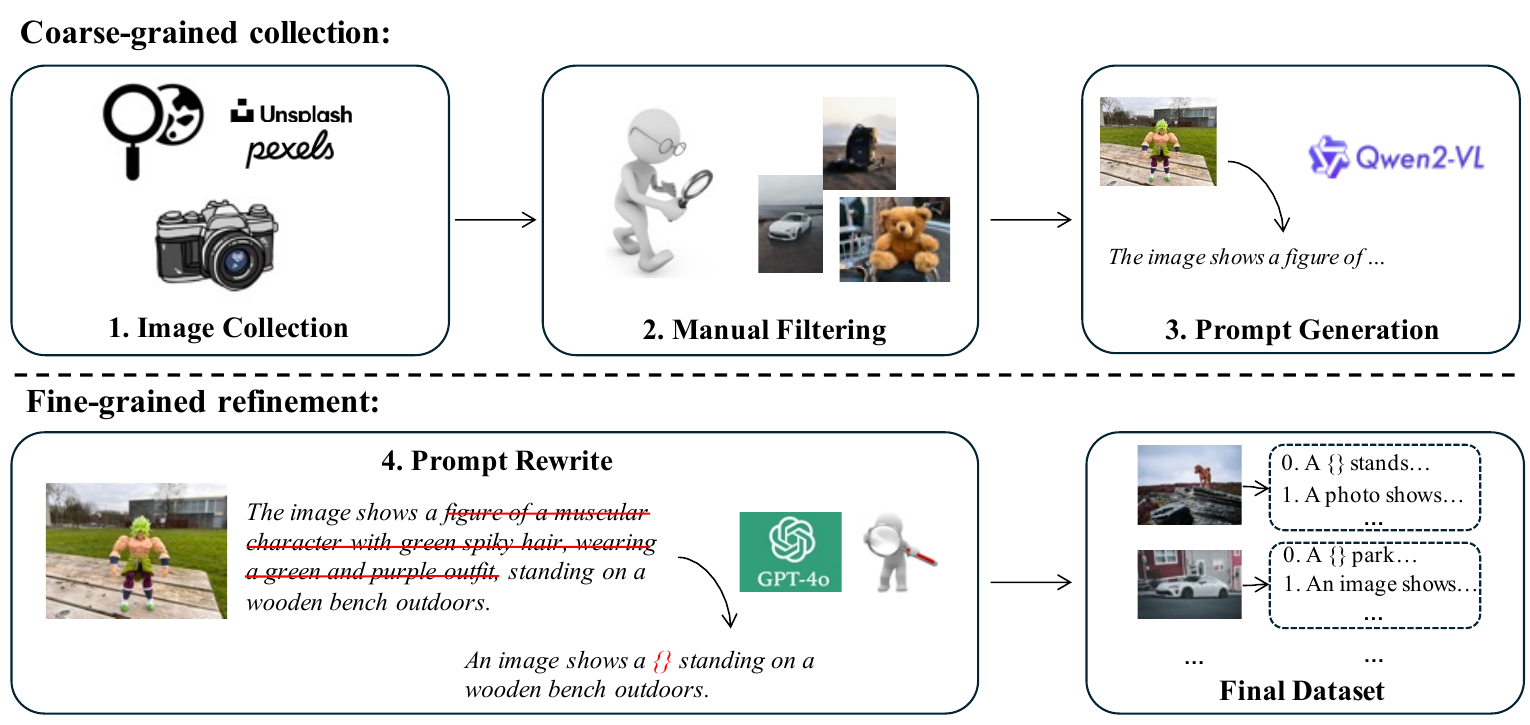}
    \caption{\textbf{Dataset Construction Process.} The process consists of image collection and caption generation. We begin by manually collecting images from Unsplash~\cite{unsplash} and Pexels~\cite{pexel}, supplemented by our own photography. To ensure high-quality data, all images undergo manual inspection and filtering by human evaluators. Prompt generation is a two-step process: first, Qwen2-VL generates initial captions for each image. These captions are then refined through human editing and GPT-4o to enhance clarity and correctness. Additionally, any text that might reveal subject-specific information is carefully removed to maintain neutrality.}
    \label{fig:dataset-pipeline}
\end{figure*}
As discussed in previous sections, the existing evaluation setup is flawed due to the absence of a separate test set, making unbiased evaluation challenging. To address this issue, we introduce \textbf{PDST}—a new dataset designed specifically for personalized image generation. Short for “\textbf{P}ersonalized \textbf{D}ataset with a \textbf{S}plit \textbf{T}est set,” \textbf{PDST} includes both a training set and a separate test set, ensuring more rigorous and reliable evaluation.

However, constructing a dataset for personalized image generation and evaluation is inherently challenging. It requires collecting multiple images of the same subject across different backgrounds and poses, ensuring sufficient diversity while maintaining subject consistency. 
To achieve this, we follow a two-step data collection process.~\cref{fig:dataset-pipeline} illustrates our dataset construction pipeline, and we provide a detailed explanation of each step below. 

\begin{figure}[t]  
    \centering
    \includegraphics[width=0.5\textwidth]{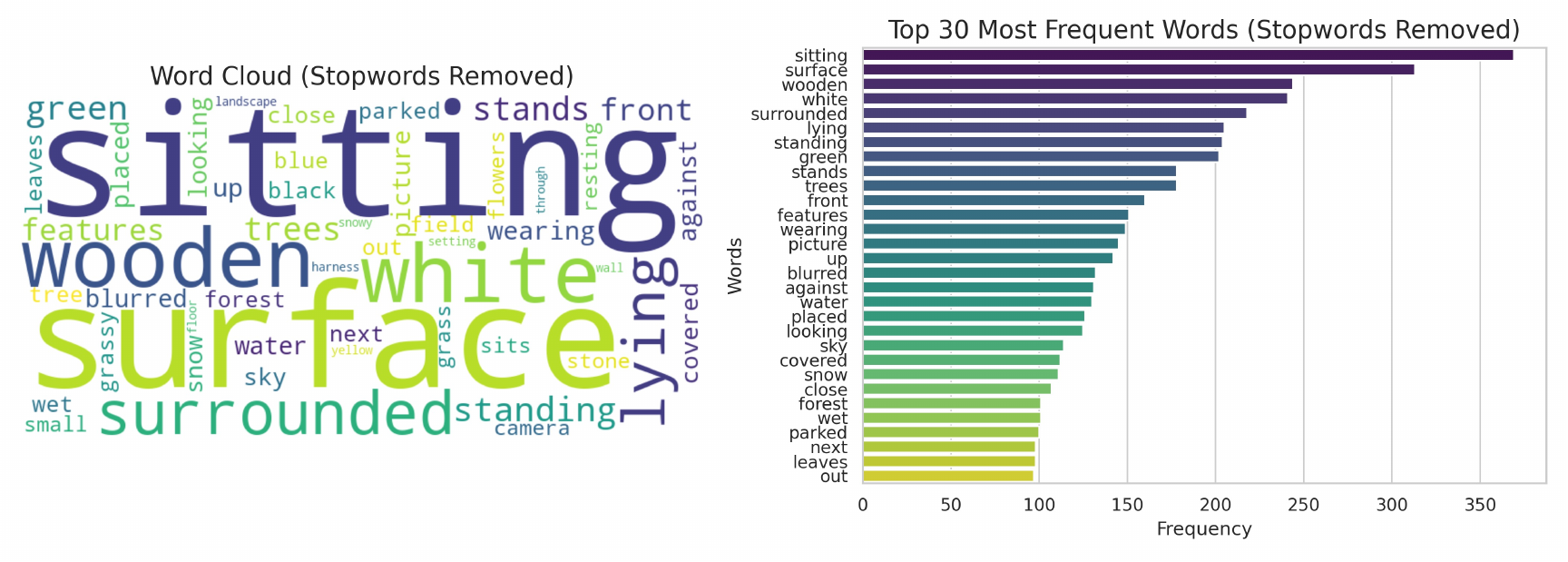}  
    \caption{Left: A word cloud visualization of the captions in the test set. Right: The top 30 most frequent words in the test set captions, with stopwords removed.}
    \label{fig:word-dis}
\end{figure}

\begin{itemize}
\item \textbf{Image Collection} We first collect web images from Unsplash~\cite{unsplash} and Pexels~\cite{pexel}, prioritizing consistency by sourcing multiple images of the same subject from the same author. To further ensure subject consistency, each set of collected images undergoes an additional verification process by two independent reviewers. If either reviewer determines that an image does not depict the same subject, it is removed from the dataset. This meticulous selection process results in $14$ subjects and each with $15$ images. To compensate for this limitation and enhance the diversity of the dataset, we also capture our own images of objects, supplementing the dataset with carefully curated, high-quality photographs.

\item \textbf{Prompt Generation}  Once the images have been collected, we utilize the powerful multimodal language model, Qwen2-VL~\cite{wang2024qwen2}, to generate 10 captions per image for the image forming the test set. The captions are then reviewed and refined by GPT-4o~\cite{hurst2024gpt} and two human reviewers. To avoid information leakage, we further remove any information related to the subject from the text and replace the subject by a \{\}. We provide more details about this process in the Appendix.

\end{itemize}
We curate the dataset containing $20$ distinct subjects, evenly divided between living entities and rigid objects ($10$ each). Each subject is represented by $15$ images, with 5 allocated for training and 10 reserved for testing, ensuring a clear separation for unbiased evaluation. Each test image is annotated with $10$ unique captions, capturing a diverse range of descriptions. The word distribution of these captions, as illustrated in~\cref{fig:word-dis}, highlights the richness and variability of our test set. For evaluation, we propose a structured process to assess subject consistency and prompt alignment. Specifically, the captions of the test images are used as prompts for the trained model to generate new images. This ensures that the generated images maintain the same semantic content as the corresponding test images, allowing for a fair and controlled comparison. 
Focused on semantic alignment rather than overfitting to training images, our approach minimizes distractions from irrelevant elements and centers on the subject itself.
In our experiments, the generated images of the test set are evaluated with automatic evaluation metrics, including the CLIP text-image similarity score, CLIP image-image similarity score, and DINO image-image similarity score, providing an objective assessment of model performance.

\section{Method}
\label{sec:method}

\subsection{Latent Text-to-Image Diffusion Model}

We implement our pipeline using a Latent Diffusion Model (LDM)~\cite{ldm}, which generates images conditioned on text in a compact latent space. The LDM consists of a pre-trained autoencoder and a learnable diffusion model. The autoencoder maps images to a latent space via an encoder $\mathcal{E}$ and reconstructs them with a decoder $\mathcal{D}$. The diffusion model, parameterized by $\theta$, learns to denoise Gaussian noise into a meaningful latent representation through iterative refinement steps.
The training objective is:

\begin{equation}
  L_{LDM} := \mathbb{E}_{\substack{
    z_0\sim \mathcal{E}(x) \\ 
    \epsilon\sim\mathcal{N}(0,1) \\ 
    t\sim \text{U}(1,T)
}} \! 
\left[
\|\epsilon - \epsilon_{\theta}(z_t, t, c(\mathcal{P})) \|_2^2
\right]
\label{eq: ldm-loss1}
\end{equation}
where $\epsilon_{\theta}$ is a learnable diffusion U-Net conditioned on a latent representation $z_t$, a timestep $t$ uniformly sampled from $(1,T)$, and a text embedding $c(\mathcal{P})$. The text embedding is obtained using a pretrained text encoder $c$, which is typically frozen during the training of the text-guided diffusion model.
During training, the model learns to remove noise and reconstruct the data distribution. During inference, it starts with random Gaussian noise and refines it into a meaningful output conditioned on the text prompt through the learned reverse process.

\subsection{Textual Inversion}
Textual Inversion~\cite{ti} is a widely used technique in personalized image generation, enabling fine-grained control over the appearance of generated images. Unlike the standard training of diffusion models, text inversion introduces a learnable token  $v^*$  that represents a specific subject. This token is initialized and optimized using a diffusion loss, effectively encoding the characteristics of a subject.
The training objective of text inversion is formulated as: 
\begin{equation}
\label{ldm-loss2}
  L_{LDM} := \mathbb{E}_{\substack{
    z_0\sim \mathcal{E}(x) \\ 
    \epsilon\sim\mathcal{N}(0,1) \\ 
    t\sim \text{U}(1,T)
}} \! 
\left[
\|\epsilon - \epsilon(z_t, t, c_{\theta}(\mathcal{P})) \|_2^2
\right]
\end{equation}
In this approach, only the token embedding of the designed  $v^*$ is finetuned, enabling the model to learn a compact and effective representation of the subject. 


\begin{figure*}
    \centering
    \includegraphics[width=0.96\textwidth]{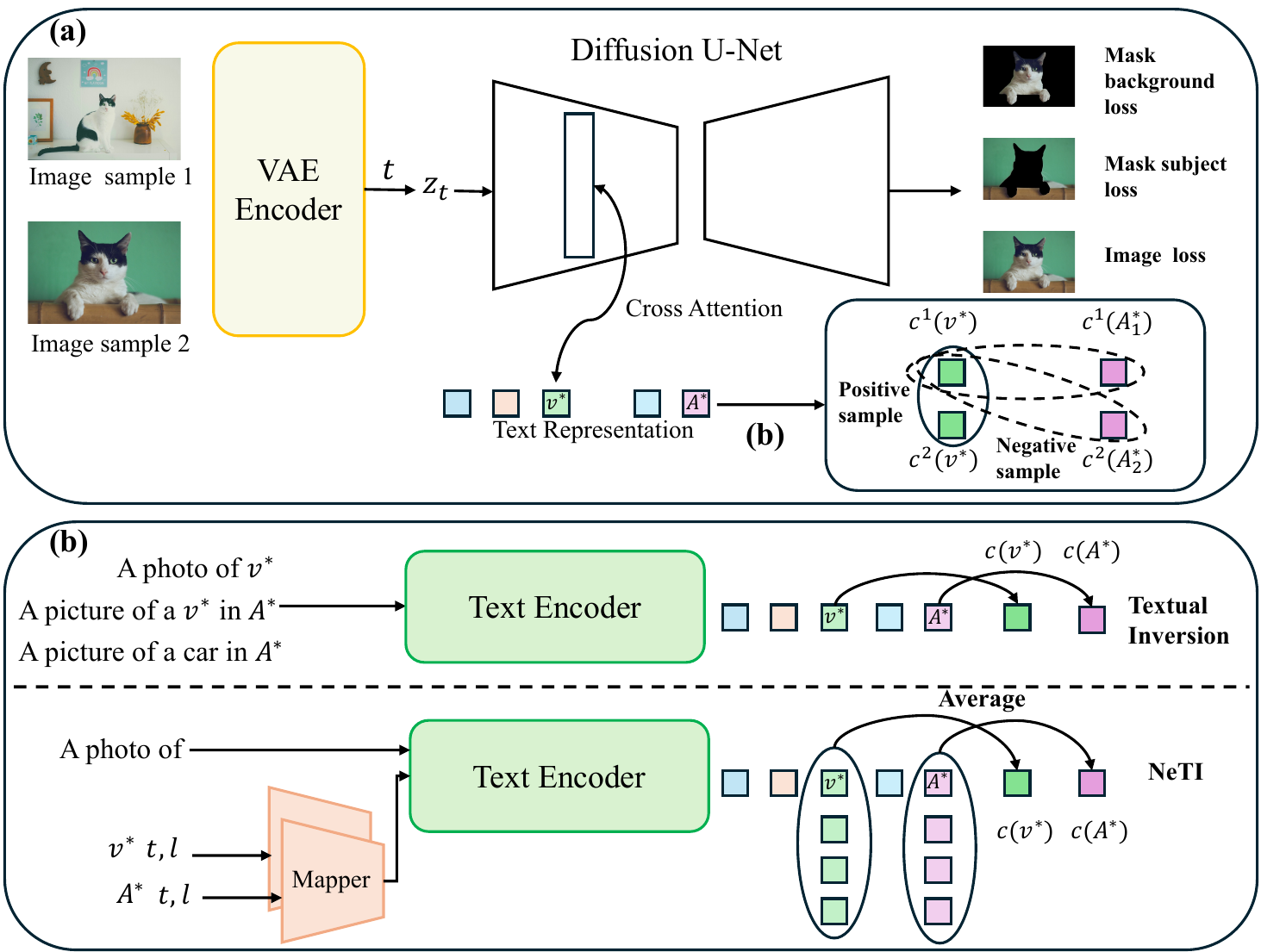}
    \caption{\textbf{Overview of our proposed training pipeline.} (a) illustrates our disentangled training losses, which include mask background loss, mask subject loss, joint loss, and contrastive loss. (b) demonstrates how we obtain learnable representations for the target subject and background attractor when applying our pipeline to Textual Inversion~\cite{ti} and NeTI~\cite{neti}. }
    \label{fig:two_column_image}
\end{figure*}

\subsection{NeTI}
The recently developed NeTI model~\cite{neti} analyzes how different layers of the diffusion U-Net contribute to image generation at various time steps. It observes that each layer plays a distinct role in shaping the final output. To capture subject-specific information, the NeTI model introduces a small neural network, $\mathcal{M}$, which takes the layer index and time index as inputs. This neural network $\mathcal{M}$ is then trained to store subject information. The training objective of NeTI can be formulated as :
\begin{equation}
    L_{NeTI}:= \mathbb{E}_{\substack{
    z_0\sim \mathcal{E}(x) \\ 
    \epsilon\sim\mathcal{N}(0,1) \\ 
    t\sim \text{U}(1,T)
}} \! 
\left[
\|\epsilon - \epsilon(z_t, t, c(\mathcal{P}),\mathcal{M}_{\theta}(t,l)) \|_2^2
\right]
\label{loss_neti}
\end{equation}
where $t,l$ are the time index and layer index, respectively. For more details about the model, we refer the reader to the original paper~\cite{neti}.
\subsection{Background Attractor}
The original Textual Inversion model~\cite{ti} solely relies on a learnable token embedding to capture subject information. While effective, this approach also inadvertently learns details from other elements of the image. A common method to mitigate this issue is masking, which selectively hides all elements except the subject. However, this can overemphasize the subject’s features, causing them to be distributed across multiple tokens. As a result, the model’s ability to follow textual prompts may be diminished.

To address these challenges, we propose a novel learning pipeline that introduces a new learnable token embedding for each image designed to capture all elements in the image except for the subject. We call this embedding the \textbf{attractor} ($A^*$ ), as it draws on background information while keeping it separate from the subject representation. To ensure effective learning, we incorporate a combination of a contrastive learning loss and a masking loss, guiding both the subject token embedding and the attractor to accurately encode their respective information. We provide a detailed explanation of our proposed pipeline below.

Given a training batch of \( N \) images featuring the same subject but with different backgrounds, along with a super-category token \( S \) representing the subject, we construct three distinct prompt pools, each designed for a specific learning objective.

\begin{itemize}
    \item \textbf{Subject Learning:} The first prompt pool consists of prompts in the format \textbf{``A photo of \( v^* \)''}, where \( v^* \) represents the subject. This helps the model focus on learning subject-specific features.

    \item \textbf{Background Learning:} The second prompt pool uses the format \textbf{``A photo of a \( S \) in the \( A^* \)''}, where \( A^* \) represents the background and \( S \) is the supercategory the subject belongs to. This prompt is designed to help the model capture background-related information.

    \item \textbf{Joint Learning:} The third prompt pool combines both subject and background information, using, for instance, the format \textbf{``A photo of a \( v^* \) in the \( A^* \)''}. This ensures that the model learns how the subject interacts with different backgrounds.
\end{itemize}

Using only three designed prompts is insufficient to ensure that different tokens learn distinct aspects of the image. 
To address this, we introduce subject and background masking losses alongside a contrastive loss, enabling tokens to capture specific information more effectively. We first use Florence~\cite{florence} to identify the subject's bounding box and then apply SAM2~\cite{ravi2024sam} for segmentation, obtaining the subject mask $M_s$ and background mask $M_b$. These masks are used to compute the loss function, formulated as follows:
\begin{equation}
\begin{aligned}
    L_{\text{sub}} := \mathbb{E}_{\substack{
    z_0\sim \mathcal{E}(x) \\ 
    \epsilon\sim\mathcal{N}(0,1) \\ 
    t\sim \text{U}(1,T)
}} \! 
\left[
\| M_s\circ\epsilon - M_s\circ\epsilon(z_t, t, C) \|_2^2
\right] \\
    L_{\text{bg}} := \mathbb{E}_{\substack{
    z_0\sim \mathcal{E}(x) \\ 
    \epsilon\sim\mathcal{N}(0,1) \\ 
    t\sim \text{U}(1,T)
}} \! 
\left[
\|\ M_b\circ\epsilon - M_b\circ\epsilon(z_t, t, C ) \|_2^2
\right]
\end{aligned}
\end{equation}
where $C$ corresponds to $c_{\theta}(\mathcal{P})$ for Textual Inversion and $c(\mathcal{P},\mathcal{M}(t,l))$ for NeTI, respectively. 
In contrast, no masking is applied in the joint learning prompts, so the loss remains unchanged as in~\cref{ldm-loss2} and~\cref{loss_neti}. This loss is denoted as $L_{joint}$.


To ensure that different tokens capture distinct aspects of the training image, we introduce a contrastive loss. In this framework, the contextual embeddings of the subject token $v^*$ (denoted as $c_{\theta}(v^*)$) from different training samples form a positive pair, 
while the embeddings of the subject token $c_{\theta}(v^*)$ and the attractor token $c_{\theta}(A^*)$ form a negative pair. The contrastive loss guides the subject embedding to learn features distinct from those captured by the attractor. This loss function is formulated as:

\begin{equation}
L_{\text{infoNCE}} = -\log \frac{\sum\limits_{(i,j) \in \mathcal{P}} e^{ v_i^{*\top} v_j^* / \tau }}
{\sum\limits_{(i,j) \in \mathcal{P}} e^{ v_i^{*\top} v_j^* / \tau } + \sum\limits_{(i,k) \in \mathcal{N}} e^{ v_i^{*\top} A_k^* / \tau }}
\end{equation}

Here,  $\mathcal{P}$ represents the set of positive pairs,  $\mathcal{N}$ denotes the set of negative pairs, and  $\tau$  is the temperature parameter. By integrating all the techniques discussed above, we formulate our final loss function as follows:

\begin{equation}
L_{\text{total}} = w_s \cdot L_{\text{sub}} + w_b \cdot L_{\text{bg}} + w_i \cdot L_{\text{joint}} + w_c \cdot L_{\text{InfoNCE}}
\end{equation}

where $w_k$ for $k \in \{s, b, i, c\}$
are weights that balance the contribution of the four losses. We present our training pipeline in~\cref{fig:two_column_image}.

\section{Experiments}
\label{sec:exp}
\textbf{Baseline and Comparison} 
Our proposed framework seamlessly integrates with existing tuning-based text-to-image personalization models.
To evaluate its effectiveness, we integrate our method into two baseline models: Textual Inversion~\cite{ti} and NeTI~\cite{neti}. We then benchmark our approach against several popular methods, including vanilla Textual Inversion, NeTI, DreamBooth~\cite{dreambooth}, and the Custom Diffusion Model~\cite{huang2024realcustom}. For a fair comparison, we retrain all models using the optimal hyperparameters reported in their respective original papers.

\noindent \textbf{Evaluation Metric}
Following previous studies~\cite{ti,dreambooth,huang2024realcustom,neti}, we evaluate the model’s prompt alignment and subject consistency. For prompt alignment, we use the CLIP text-image similarity score~\cite{dreambooth}, 
For subject consistency, we employ the CLIP image-image similarity score~\cite{dreambooth} and the DINO image-image similarity score~\cite{dreambooth}. Unlike previous works that compute these scores using training data, we evaluate them using our proposed separate test set. This test set consists of $20$ subjects with each subject containing 10 images and each image accompanied by $10$ descriptive sentences. During inference, we generate $5$ images per sentence, resulting in a total of $10,000$ generated images per model. The final score for each metric is obtained by averaging the scores of all generated images.
\begin{figure}[t]  
    \centering
    \includegraphics[width=0.5\textwidth]{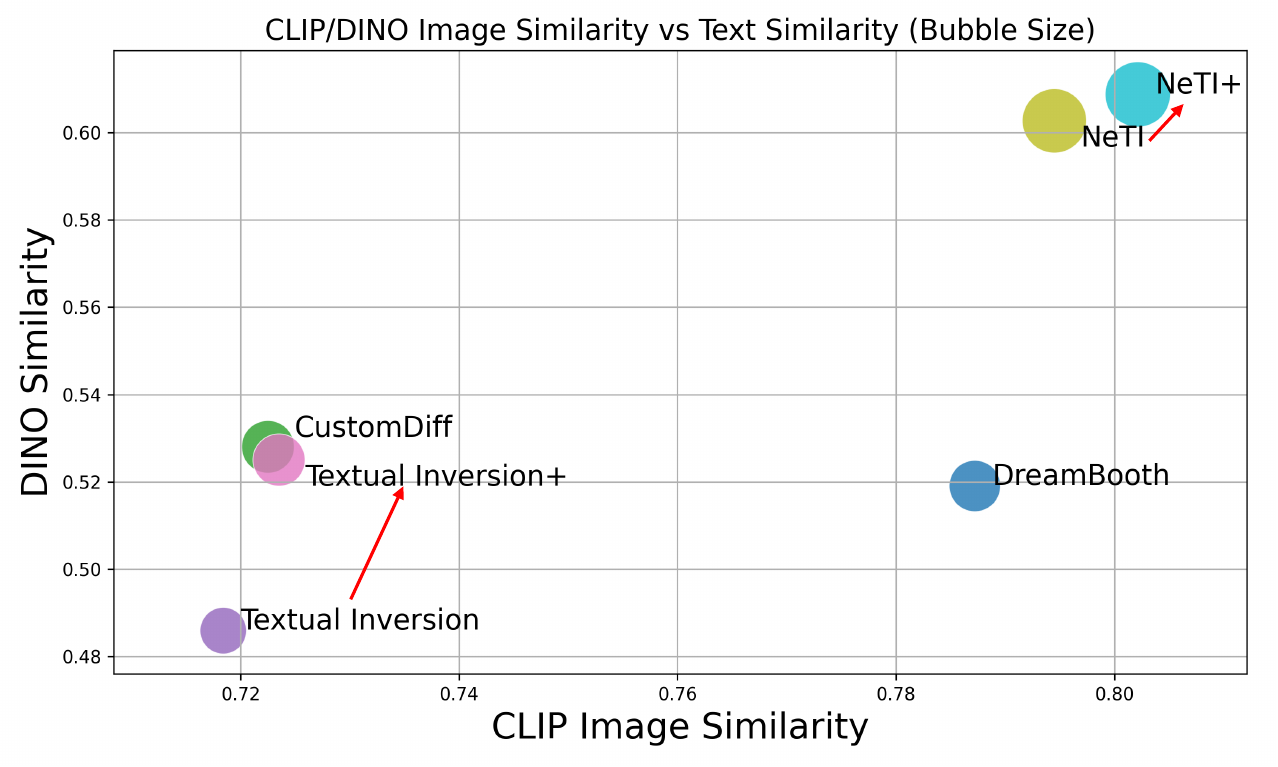}  
    \caption{Quantitative evaluation: Comparing CLIP/DINO image similarity versus text similarity, with bubble sizes indicating text similarity scores, where a larger size corresponds to a better score.}
    \label{fig:result}
\end{figure}

\noindent \textbf{Implementation Details}
We implement all methods using the open-source Stable Diffusion 1.5~\cite{tang2023stable} and adapt our proposed training pipeline to both Textual Inversion, a widely used fine-tuning technique, and NeTI, an advanced method for subject-driven generation. For training, we employ the AdamW optimizer~\cite{adamw} with a learning rate of $1e-5$ and a batch size of $8$. To maintain stable training dynamics, we use a constant learning rate scheduler. This setup is implemented in both methods.
\section{Results}
\begin{figure*}[htbp]
    \centering
    \includegraphics[width=1.0\textwidth]{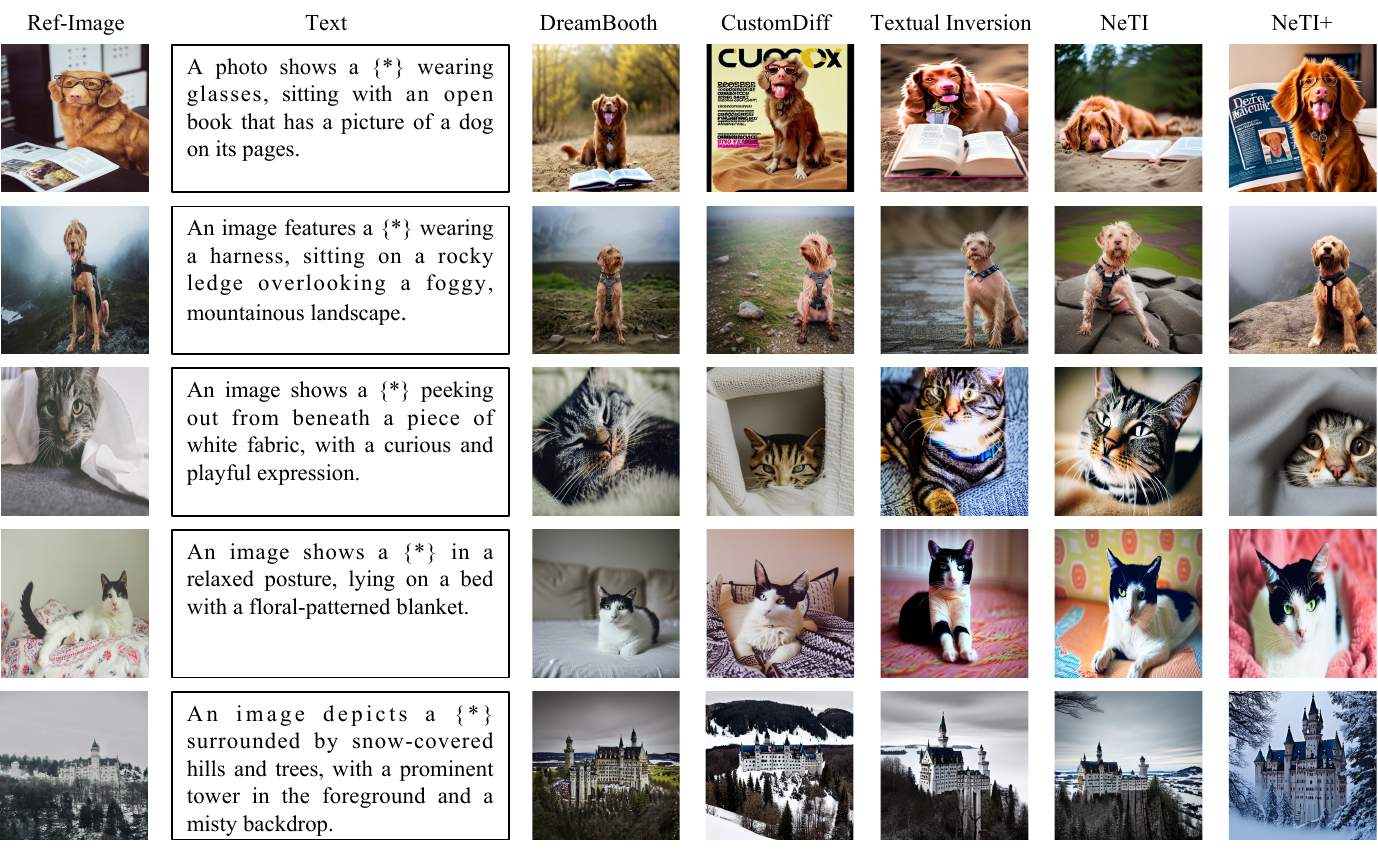}
    \caption{\textbf{Qualitative Comparison.} We use the same prompt to guide all models in generating images. These prompts are derived from the reference images in the test set, which are shown in the first column.}
    \label{fig:vis1}
\end{figure*}
We present a quantitative comparison of results in~\cref{fig:result}. The $x$- and $y$-axes represent the DINO and CLIP image-image similarity scores, respectively, while the bubble size reflects the CLIP text-image similarity score, offering a comprehensive view of model performance across multiple dimensions.
Among the base models, NeTI~\cite{neti} achieves the highest performance across all evaluation metrics, consistent with findings from prior studies. To further enhance model effectiveness, we apply our proposed training pipeline to both Textual Inversion~\cite{ti} and NeTI, referring to the improved versions as Textual Inversion$+$ and NeTI$+$, respectively.
As shown in~\cref{fig:result}, our training pipeline consistently improves the performance of both models across all metrics. This improvement is visually emphasized by the red arrows, underscoring the effectiveness of our approach in enhancing model performance.

In~\cref{fig:vis1}, we provide a visual comparison of all models. The first column shows reference images from the test set, used to generate the textual descriptions in the second column. The remaining columns display images generated by different models based on these prompts. Our proposed training method improves the performance of both Textual Inversion and NeTI.
Given that NeTI serves as a significantly stronger baseline, integrating our method with it results in the best overall performance (NeTI$+$).
For clarity, we primarily focus on NeTI$+$ results and omit those from Textual Inversion$+$.
As seen in the figure, all models generate plausible images, and due to the detailed prompts, the outputs share common semantic elements. However, NeTI$+$ consistently produces images with superior subject consistency and alignment with the text. For instance, in the first row, only NeTI$+$ accurately captures the detail “a picture of a dog on its page.” Similarly, in the third row, NeTI$+$ uniquely generates an image of a cat with a curious and playful expression, matching the prompt. 
See the Appendix for more qualitative results and comparisons.

\begin{figure}[t]  
    \centering
    \includegraphics[width=0.49\textwidth]{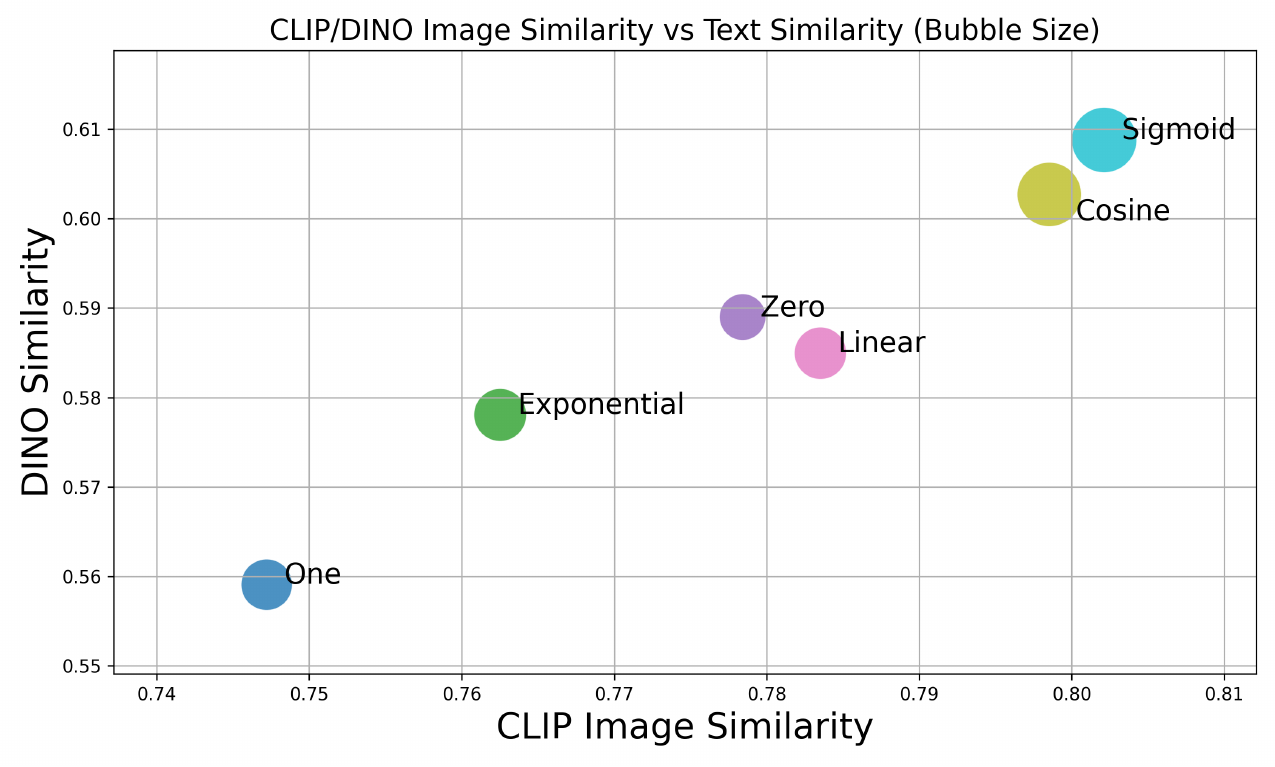}  
    \caption{Performance of models trained with different contrastive loss weighting functions. Zero and One refer to the identity function with value zero and one, respectively.}
    \label{fig:weight-per}
\end{figure}

To highlight the improved capabilities of NeTI$+$, we provide a visual comparison between NeTI and NeTI$+$ using prompts with fictional designs in~\cref{fig:vis2} (Appendix). As shown, NeTI$+$ demonstrates a superior ability to interpret and capture information about fictional designs from text, whereas the standard NeTI model exhibits overfitting, generating images strongly similar to training data. This overfitting limits NeTI's generalization and accurate understanding of textual descriptions.

\begin{figure*}[htbp]
    \centering
    \includegraphics[width=1.0\textwidth]{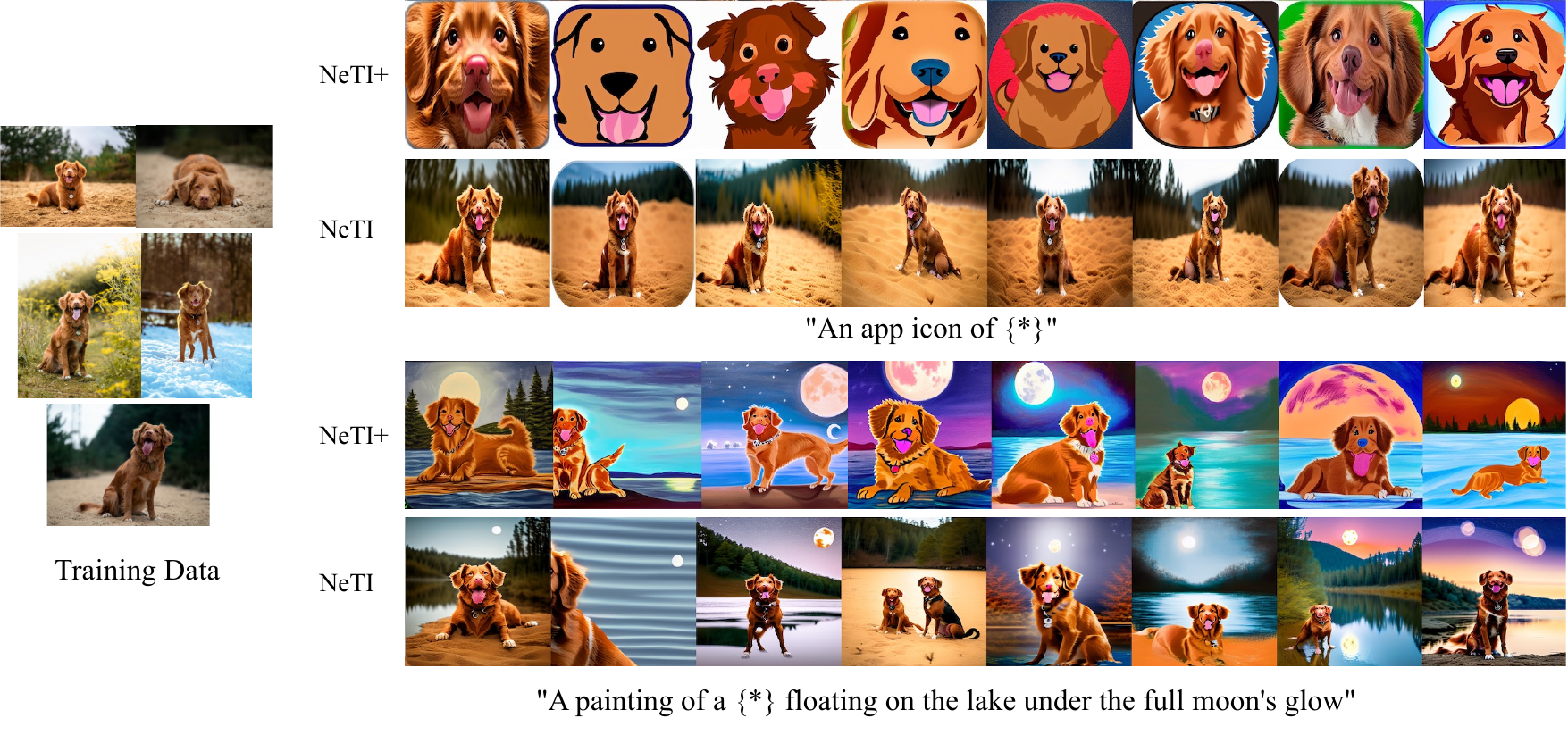}    \caption{\textbf{Qualitative Comparison between NeTI and our NeTI$+$.} On the left side, we present the training images. For each model, we generate eight images using the same prompt and random seeds.}
    \label{fig:vis2}
\end{figure*}
\subsection{Ablation Study}

\begin{figure}[tbp]  
    \centering
    \includegraphics[width=0.5\textwidth]{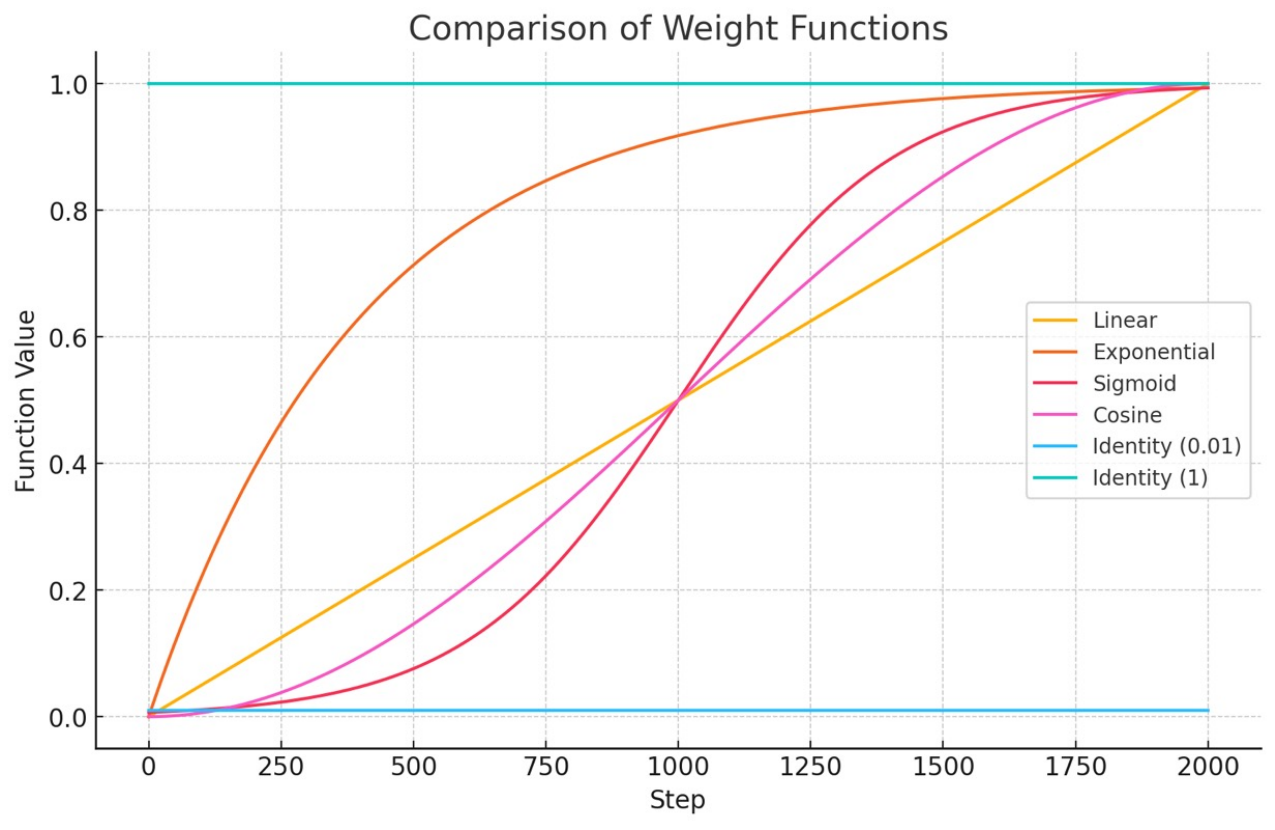}  
    \caption{Weighting function for contrastive loss. The $x$-axis corresponds to the training steps.}
    \label{fig:w-func}
\end{figure}

\textbf{Contrastive Loss} 
We first analyze the contrastive loss to disentangle the learning of subject and background. In this experiment, we explore different weighting functions for the contrastive loss, including exponential, linear, sigmoid, and cosine functions, as well as identity functions. Notably, the identity function with a constant value of zero corresponds to training without contrastive loss. This investigation helps us to understand the impact of various weighting strategies on the learning process. We present the plots of these weighting functions in~\cref{fig:w-func} (Appendix).~\cref{fig:weight-per} showcases the performance of the NeTI$+$ model trained with each approach. Our results reveal that the cosine and sigmoid weighting functions lead to superior performance compared to the others. In contrast, the exponential weighting function and the identity function with a constant value of one perform worse than training without the contrastive loss, as demonstrated by the identity function (value = 1) in~\cref{fig:weight-per}. These findings suggest that assigning a low weight to the contrastive loss at the beginning of training is crucial. This allows the dedicated embeddings to first learn both subject and background information before the contrastive loss is introduced to refine their separation. By gradually increasing the weight of the contrastive loss, we ensure that each embedding focuses on learning its respective information in the image effectively, leading to better result.

\begin{figure}[t]  
    \centering
    \includegraphics[width=0.5\textwidth]{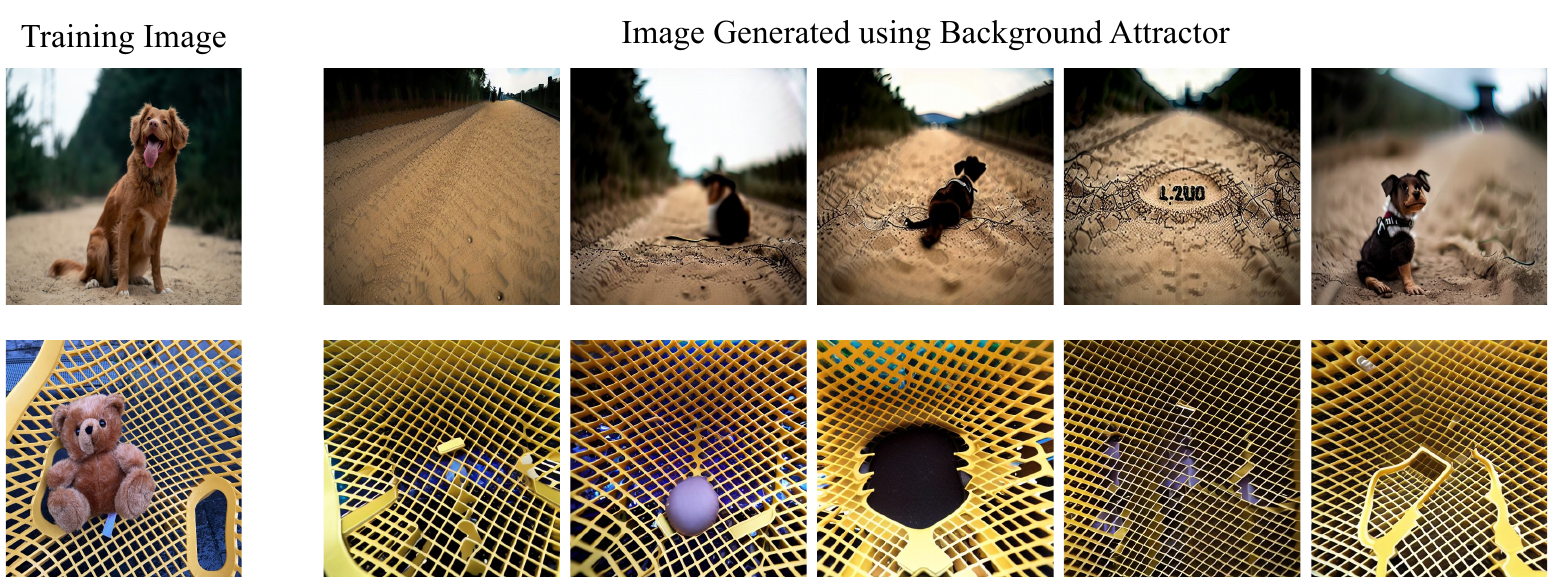}  
    \caption{Images generated using the background attractor with the prompt template: ``A photo of \{\}.’’}
    \label{fig:bg}
\end{figure}

\noindent \textbf{Background Attractor} 
Next, we analyze the information captured by our proposed background attractor. Using the prompt template ``A photo of {}.’’ with five different random seeds, we generate images, shown in~\cref{fig:bg}. The results demonstrate that the background attractor effectively learns and captures the overall characteristics of the image background, though not reproducing it exactly. This indicates successful encoding of background information.

\section{Conclusion}
\label{sec:con}
In this work, we introduced a new dataset for the image personalization task. Unlike previous datasets that only provide training images, our dataset includes a separate test set, enabling unbiased evaluation. Additionally, we proposed a novel training pipeline that incorporates a background attractor and weighted contrastive loss to effectively disentangle the subject from the background. This approach ensures that the subject embedding focuses on learning the subject itself without being influenced by other elements in the images. Our proposed training pipeline can be directly applied to Text Inversion~\cite{ti} and NeTI~\cite{neti} methods. Compared to previous approaches, models trained with our pipeline demonstrate a superior ability to capture subject details while also exhibiting stronger text understanding. These improvements highlight the effectiveness of our approach in advancing image personalization techniques.

\section{Limitations}
Our dataset has some limitations. First, our dataset does not include samples for style subject learning, which limits its applicability in stylistic personalization. Second, the test set currently focuses only on single-subject evaluation, restricting its ability to assess multi-subject scenarios. In the future, we plan to expand our dataset to address these limitations by incorporating samples for style learning and introducing a test set that supports multi-subject evaluation.

A limitation of our method is its inability to learn style subjects. While we disentangle subject and background using a mask and a contrastive loss, style is an abstract property that cannot be separated this way. However, our contrastive loss approach could potentially be adapted to disentangle image content from style, which we aim to explore in future work.

{
    \small
    \bibliographystyle{ieeenat_fullname}
    \bibliography{main}
}

\clearpage
\setcounter{page}{1}
\maketitlesupplementary
\section{Social Impact}
Image personalization offers both exciting opportunities and ethical challenges. On the positive side, it enables users to generate highly tailored visuals, fostering creativity, accessibility, and personalization in marketing, entertainment and design. It can also help individuals see themselves represented in media more accurately. However, personalization raise concerns about misinformation, privacy and bias. More realistic personalized images could be misused for deepfakes, identity fraud, or spreading false narratives. Bias in training data may also lead to unfair or exclusionary representations. Balancing innovation with ethical safeguards is crucial to ensuring image personalization enhances creativity without enabling harm.

\section{Dataset}
\begin{figure*}[htbp]
    \centering
    \includegraphics[width=1.0\textwidth]{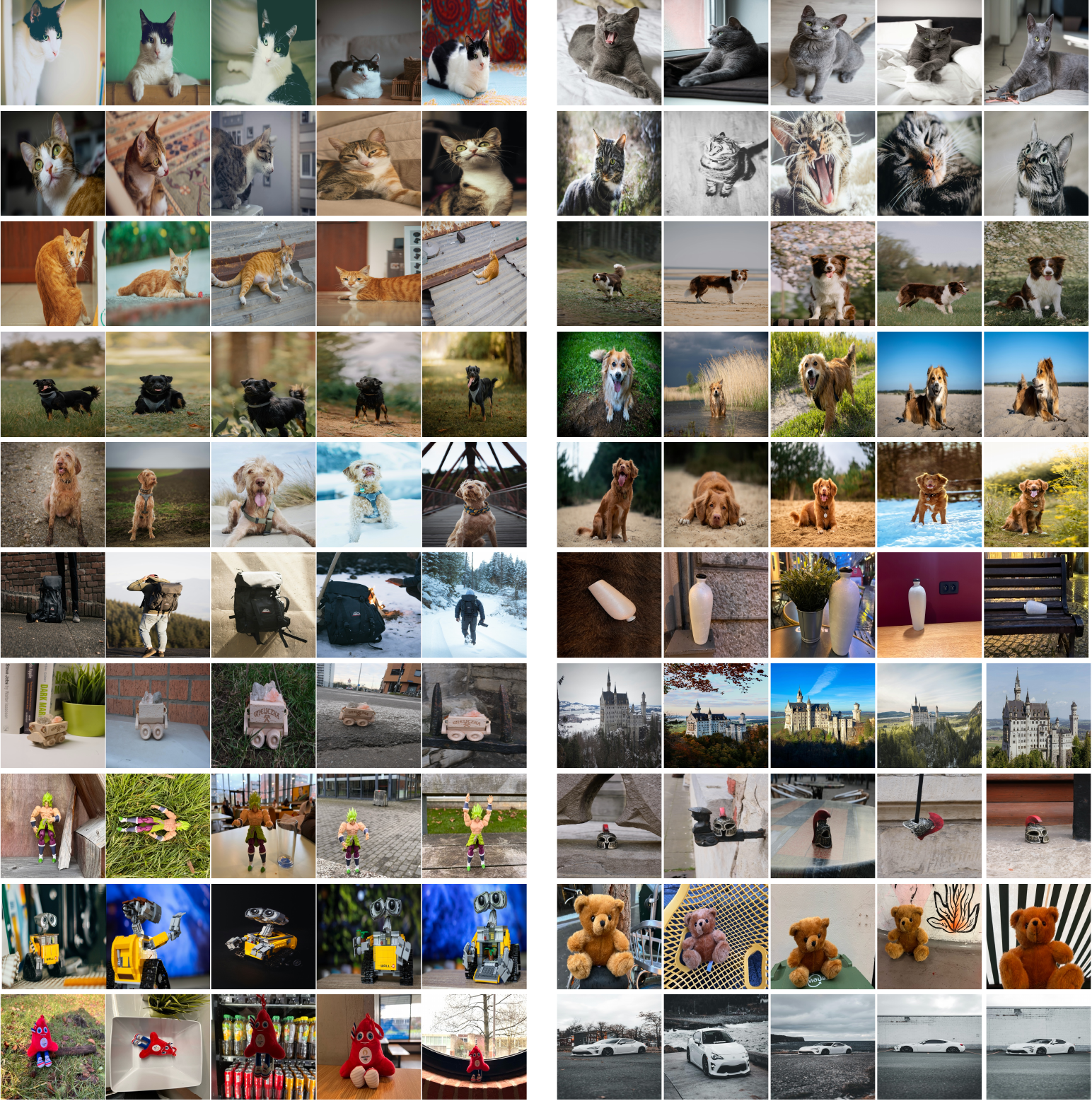}    \caption{Training Image. Our dataset contains $20$ subjects, and each subject has $5$ images for training.}
    \label{fig:appendix-vis1}
\end{figure*}

Below are prompt templates that are used during model training: 

\textbf{Prompts used during training.} 
\begin{itemize}
    \item ``a photo of a \{\}.''
    \item ``a rendering of a \{\}.''
    \item ``the photo of a \{\}.''
    \item ``a photo of a clean \{\}.''
    \item ``a photo of a dirty \{\}.''
    \item ``a dark photo of the \{\}.''
    \item ``a photo of the cool \{\}.''
    \item ``a close-up photo of a \{\}.''
    \item ``a bright photo of the \{\}.''
    \item ``a cropped photo of a \{\}.''
    \item ``a photo of the \{\}.''
    \item ``a good photo of the \{\}.''
    \item ``a close-up photo of the \{\}.''
    \item ``a rendition of the \{\}.''
    \item ``a photo of a nice \{\}.''
    \item ``a photo of a \{\} in the \{\}.''
    \item ``a rendering of a \{\} in the \{\}.''
    \item ``a cropped photo of the {} in the {}.''
    \item ``the photo of a \{\} in the \{\}.''
    \item ``a photo of a clean \{\} in the \{\}.''
    \item ``a photo of my \{\} in the \{\}.''
    \item ``a photo of the nice \{\} in the \{\}.''
    \item ``a good photo of a \{\} in the \{\}.''
    \item ``a rendition of a \{\} in the \{\}.''
    \item ``a photo of the clean \{\} in the \{\}.''
    \item ``a photo of a cool \{\} in the \{\}.''
    \item ``a close-up photo of a \{\} in the \{\}.''
    \item ``a photo of the cool \{\} in the \{\}.''
    \item ``a cropped photo of a \{\} in the \{\}.''
    \item ``a photo of one \{\} in the \{\}.''
\end{itemize}
~\cref{fig:appendix-vis1} shows the training data, which contains $20$ different subjects, and each subject has $5$ different images.
\begin{figure}[t]  
    \centering
    \includegraphics[width=0.5\textwidth]{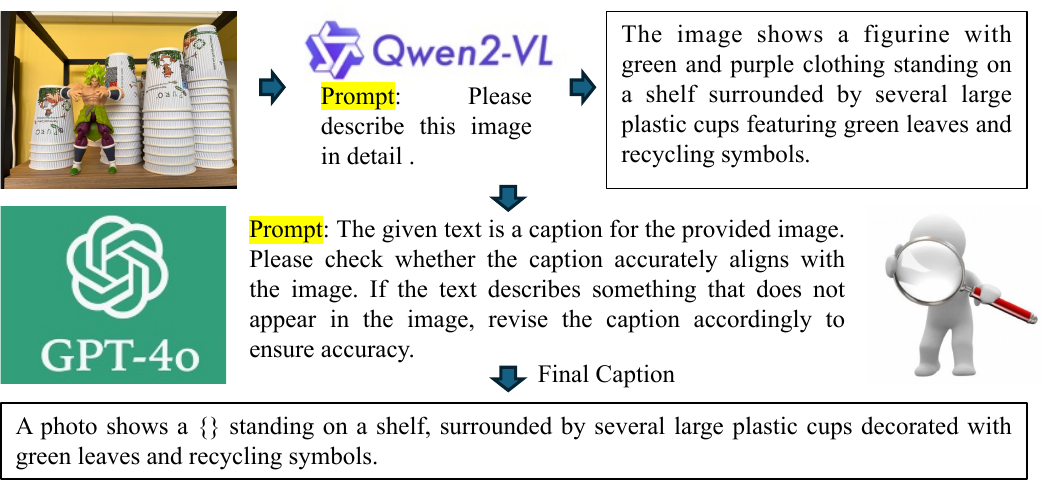}  
    \caption{The pipeline of generating high-quality captions for images in test set.}
    \label{fig:cap-gen}
\end{figure}
During our data construction, we first use QWen2-VL~\cite{wang2024qwen2} to generate initial prompts. These prompts are then reviewed and refined by GPT-4o to improve clarity and coherence. To further ensure that no subject-specific information is unintentionally included, we conduct a final manual review, carefully removing any such details. This multi-step process helps maintain the neutrality and generality of the prompts. A visual representation of this workflow is provided in~\cref{fig:cap-gen}.

\section{Evaluation Setting}
\begin{figure*}[htbp]
    \centering
    \includegraphics[width=1.0\textwidth]{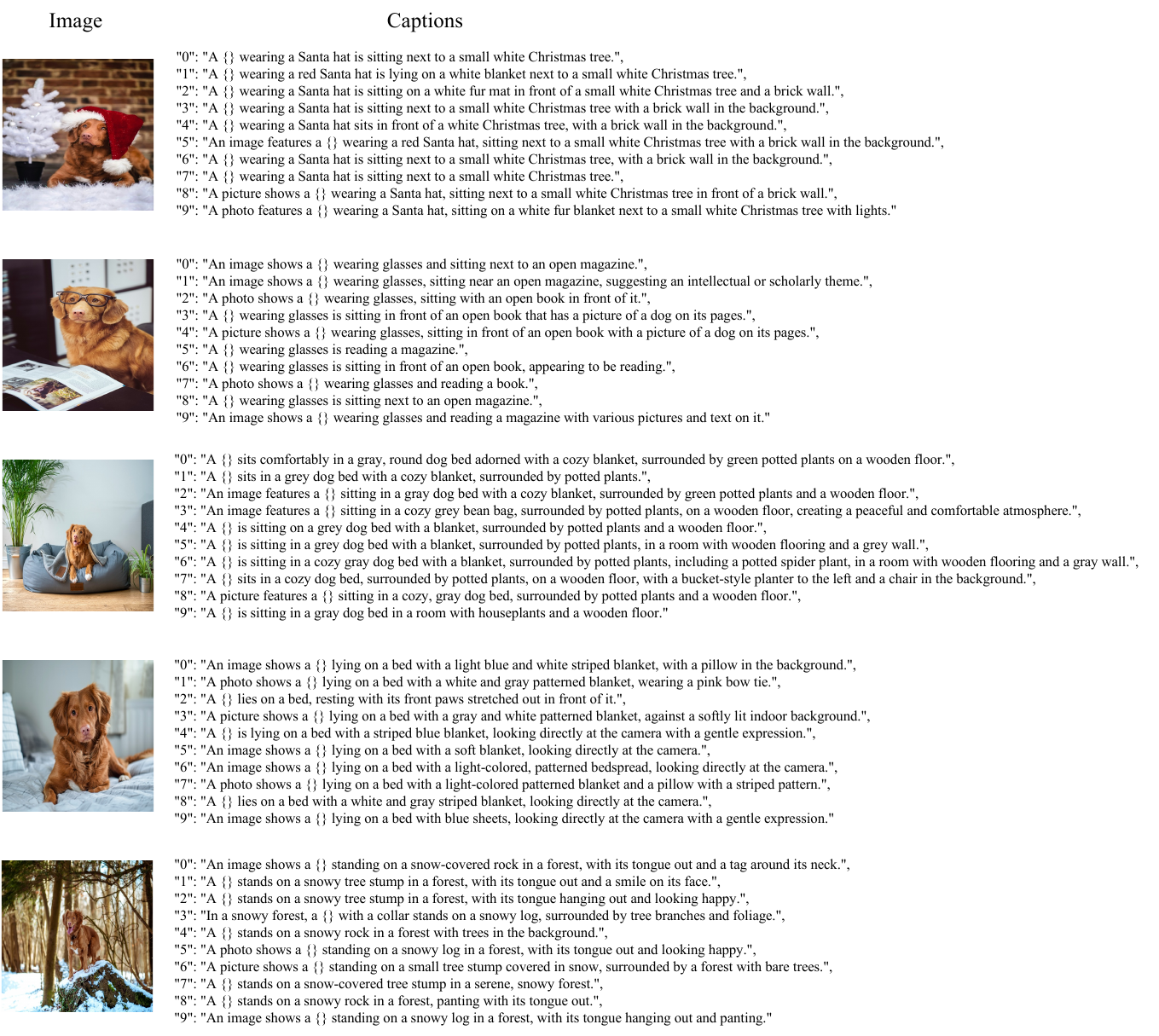}    \caption{Sample of Test Set. We present five data samples from the test set. Each image is associated with $10$ different prompts, which describe various details of the image, excluding the subject.}
    \label{fig:appendix-vis2}
\end{figure*}

In this section, we detail the evaluation process on a separate test set. For each subject in the training set, we provide $10$ test images. Each test image is paired with $10$ different captions, describing various aspects of the image while explicitly excluding the subject. An example from the test set is shown in~\cref{fig:appendix-vis2}. During testing, each caption serves as a prompt to guide the trained model in generating 10 images, each with a different random seed. Once the images are generated, we compute the CLIP space text-image similarity score between the prompt and the generated images. Additionally, to evaluate image-image similarity in both CLIP space and DINO space, we compare the generated images against the reference image that was originally used to create the prompt. We then take the average as the final score for both text similarity score and image similarity score. 

\section{Additional Results}
\begin{figure*}[htbp]
    \centering
    \includegraphics[width=1.0\textwidth]{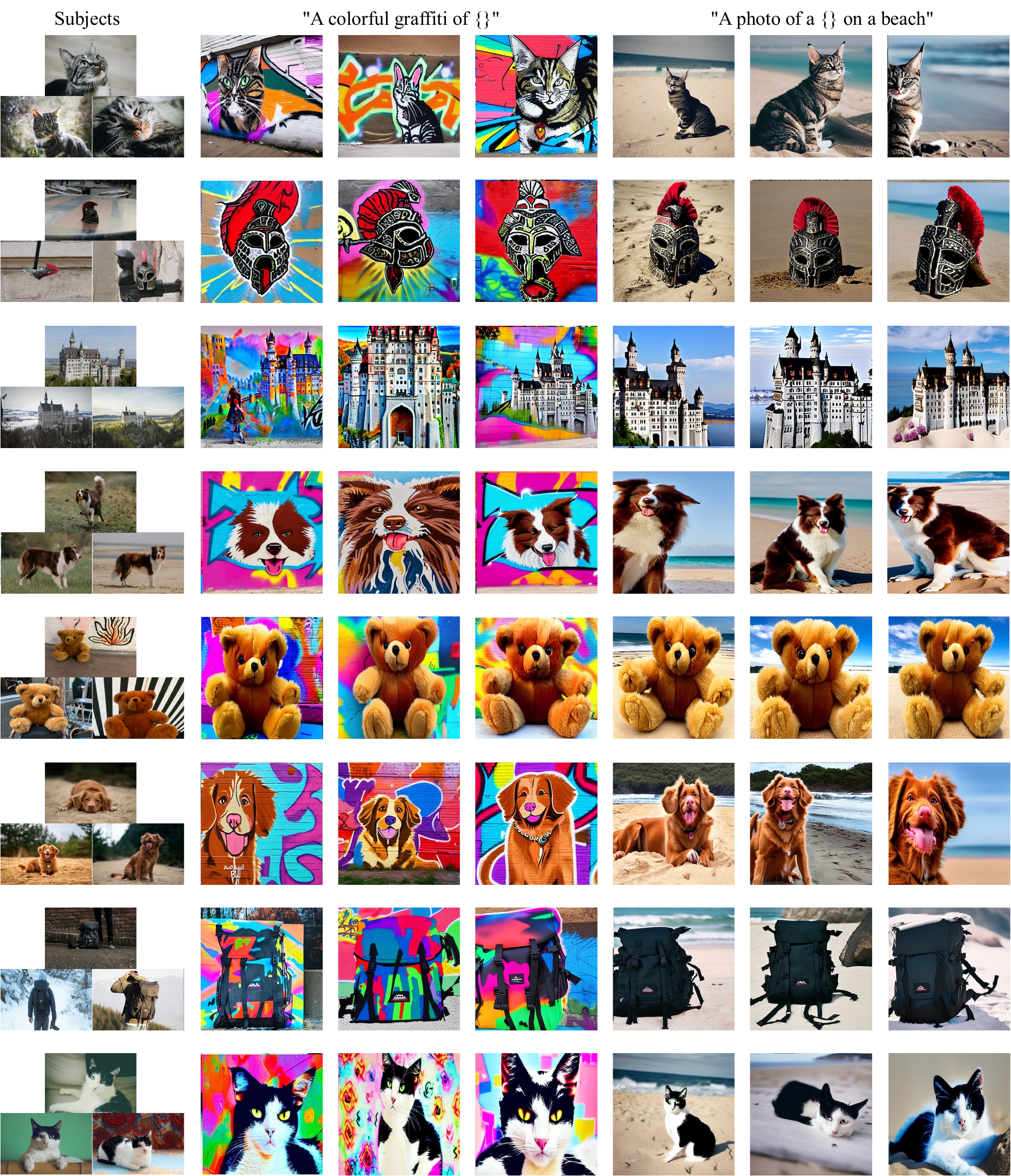}    \caption{Images generated using the NeTI$+$ model. The first column shows the training image for each subject, and for each prompt, three images are generated with different seeds.}
    \label{fig:more-vis}
\end{figure*}

\cref{fig:more-vis} presents additional images generated using the NeTI$+$ model. We use the prompts A colorful graffiti of \{\}.'' and A photo of a {} on a beach.’’ to generate three images with different seeds for each subject shown in the first column. The first prompt evaluates the model’s ability to understand and combines fictional designs, while the second assesses its capability to generate a subject within a given background. As shown in the results, the model not only accurately interprets the prompts but also maintains strong subject consistency.

\begin{figure*}[htbp]
    \centering
    \includegraphics[width=1.0\textwidth]{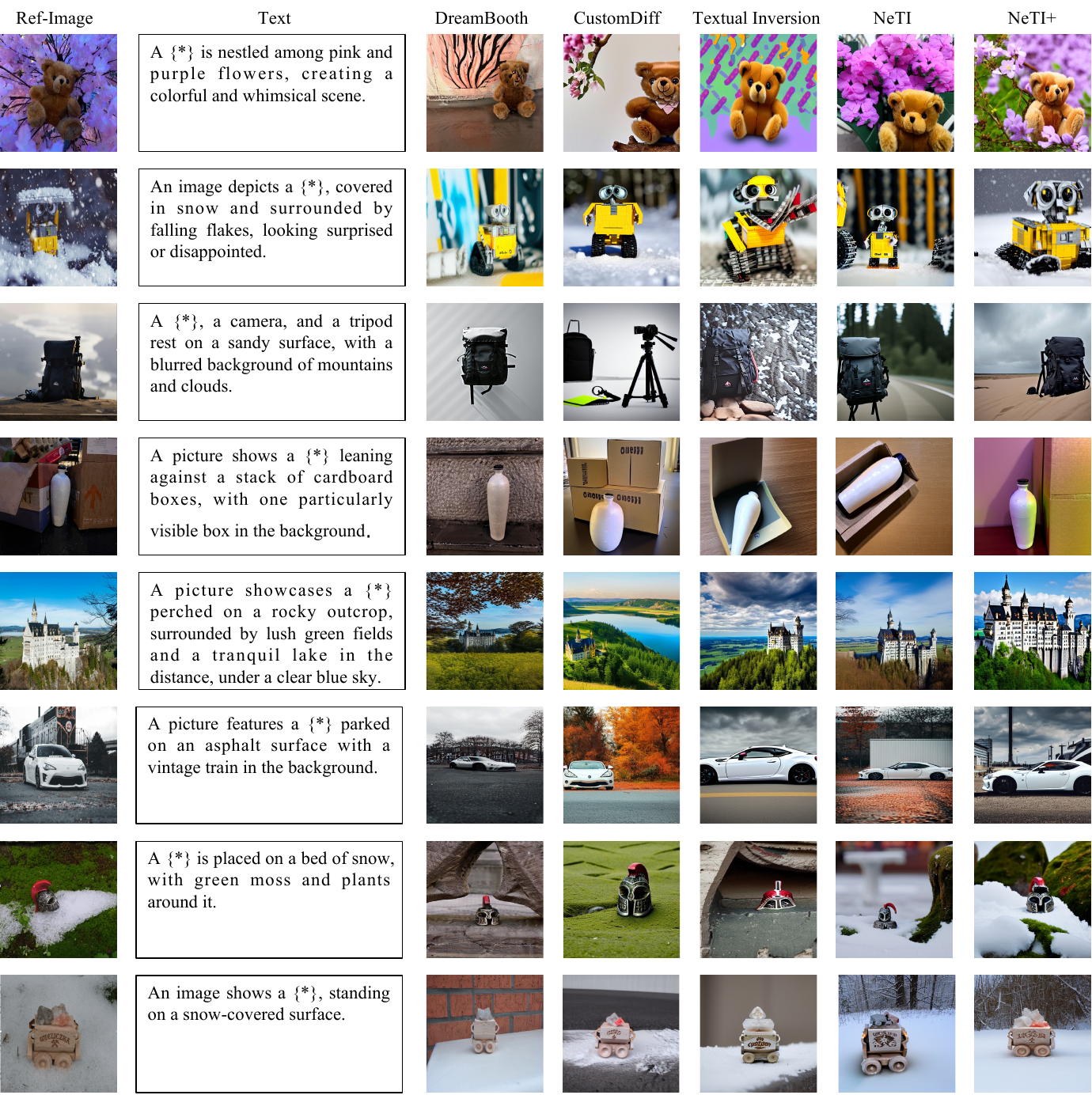}    \caption{ \textbf{Qualitative Comparison.} We use the same prompt to guide all models in generating images. These prompts are derived from the reference images in the test set, which are shown in the first column.}
    \label{fig:vis3}
\end{figure*}

Additional comparisons are shown in~\cref{fig:vis2} and~\cref{fig:vis3}, demonstrating that the NeTI$+$ model excels in generating text-aligned and subject-consistent images.

\begin{figure*}[htbp]
    \centering
    \includegraphics[width=1.0\textwidth]{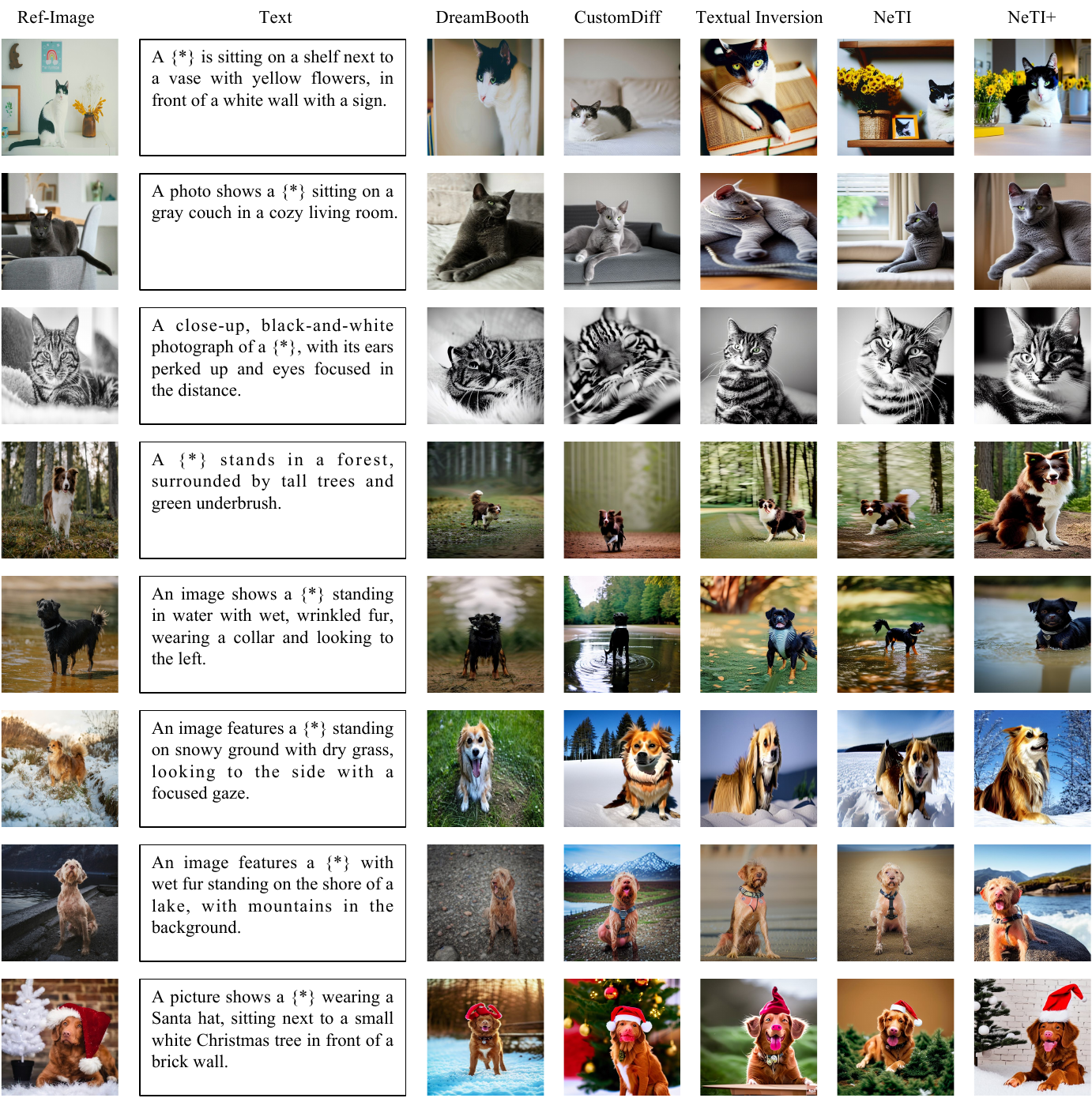}    \caption{ \textbf{Qualitative Comparison.} We use the same prompt to guide all models in generating images. These prompts are derived from the reference images in the test set, which are shown in the first column.}
    \label{fig:vis2}
\end{figure*}

\end{document}